\documentclass[10pt,twocolumn,letterpaper]{article}

\usepackage{spconf}
\usepackage{times}
\usepackage{epsfig}
\usepackage{graphicx}
\usepackage{amsmath}
\usepackage{amssymb}
\usepackage{xcolor}
\usepackage{url}
\usepackage[linesnumbered,ruled,vlined]{algorithm2e}
\usepackage[noend]{algpseudocode}
\usepackage{comment}
\usepackage{caption}

\definecolor{vyellow}{rgb}{0.7490,0.5647,0}
\definecolor{vyelloworig}{rgb}{1,0.7529,0}
\definecolor{vmagenta}{rgb}{0.4392,0.1882,0.6274}
\definecolor{vmagenta3}{rgb}{0.7254,0.6352,0.7960}
\definecolor{vmagentaorig}{rgb}{0.9215,0.8039,1}
\definecolor{vblue}{rgb}{0.3686,0.7921,0.8431}
\definecolor{vred}{rgb}{0.7529,0,0}

\newcommand{\eg}{e.g.\@\xspace} 

\newcommand{\ie}{i.e.\@\xspace}

\newcommand{\net}{ST-VAE\@\xspace}
\newcommand{\subnet}{VLT\@\xspace}

\renewcommand{\paragraph}[1]{\vspace{1mm} \noindent \textbf{#1}}

\usepackage[pagebackref=true,breaklinks=true,colorlinks,bookmarks=false]{hyperref}

\begin{document}

%%%%%%%%% TITLE
\title{Multiple Style Transfer via Variational AutoEncoder}

\name{Zhi-Song Liu, Vicky Kalogeiton and Marie-Paule Cani}
\address{LIX, École Polytechnique, CNRS, IP Paris\\
{\small{\url{https://www.lix.polytechnique.fr/geovic/project-pages/icip-style-transfer}}}
}

\maketitle
%\thispagestyle{empty}

%%%%%%%%% ABSTRACT
\begin{abstract}
Modern works on style transfer focus on transferring style from a single image. 
Recently, some approaches study multiple style transfer; these, however, are either too slow or fail to mix multiple styles. We propose \net, a Variational AutoEncoder for latent space-based style transfer. 
It performs multiple style transfer by projecting nonlinear styles to a linear latent space, enabling to merge styles via linear interpolation 
before transferring the new style to the content image. 
To evaluate \net, we experiment on COCO for single and multiple style transfer. We also present a case study revealing that \net outperforms other methods while being faster, flexible, and setting a new path for multiple style transfer.

\end{abstract}

\section{Introduction}
Style transfer is a well-visited topic~\cite{style_1,style_3,AdaIN, WCT}. 
Given a target and a reference image, its goal is to synthesize an image with content from the target and style from the reference. 
Typically, `style' refers to color, texture, or brushstroke~\cite{cycleGAN,WikiArt}. 

Most style transfer works are limited as they typically focus either \emph{solely} on single style~\cite{style_1,style_3,WCT,style_5} or on \emph{one set} of styles~\cite{cycleGAN,multi_1,multi_2}.
In real-life applications, however, a user is rarely satisfied with a single style and instead seeks for multiple ones usually by iteratively applying styles~\cite{Gatys}. 
To address multiple style transfer, the common practice is to interpolate different styles or assign different weights in the feature space~\cite{style_3,AdaIN}. Besides slow, the main limitation of such methods is that features are nonlinear and linear interpolation cannot guarantee style mixture in spatial space. 

To tackle these issues, we introduce a Variational AutoEncoder for latent space-based style transfer, coined \net.  
It is a flexible framework that adapts to single or multiple style transfer. 
It consists of (1) an Image AutoEncoder for image reconstruction, where the style manipulation takes place in the feature space instead of the pixel space (Section~\ref{sub:iae}); and (2) a Variational autoencoder-based Linear Transformation (\subnet) that first learns the feature covariance for style and content images (Section~\ref{subsub:lt}) and then maps the covariance to a latent space via KL divergence (Section~\ref{subsub:vae}). 
Multiple style transfer is achieved as latent space-based linear interpolation. 
Experiments on the COCO dataset~\cite{COCO} and comparisons to modern methods show that \net achieves fine visual quality (Section~\ref{sec:experiments}). We also conduct a user study revealing that our results are superior to the state of the art.

Our contributions are: 
(1) we introduce \net, a novel method for single and multiple style transfer; 
(2) it casts style fusion as mixture models setting the path for future study; and 
(3) it outperforms all methods quantitatively and qualitatively.

\begin{figure*}[ht!]
	\begin{center}
		\centerline{\includegraphics[width=0.96\textwidth]{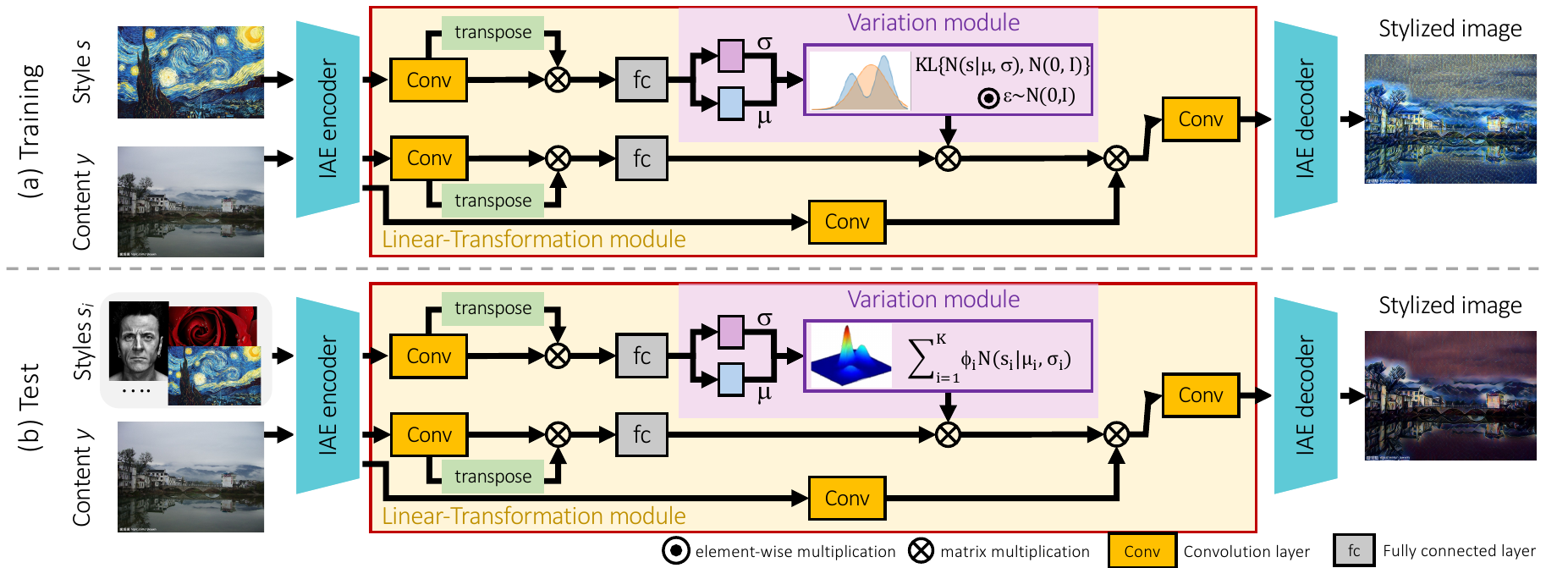}}
		\caption{\small{\net. (a) At training, \textcolor{vblue}{\textbf{IAE}} is trained and then is kept frozen. After, \textcolor{vred}{\textbf{\subnet}} learns the covariance of style images via \textcolor{vyellow}{\textbf{linear-transformation module}} and projects them to components of a multi-dimensional Gaussian distribution via KL minimization via the \textcolor{vmagenta}{\textbf{variation module}}. To transfer styles, it computes the covariance between style and content and multiplies it to the content features to obtain the new projected features.
		(b) At test time, given several input style images, \net samples styles in the latent space by model mixture (\textcolor{vmagenta}{\textbf{magenta}}).  
		}}
		\label{fig:Figure1}
	\end{center}
\end{figure*}

\section{RELATED WORK}

\paragraph{Single image style transfer. } 
Style transfer with DL was introduced in~\cite{Gatys}. 
Since then, it has gained a lot of popularity and several works address it by using features to blend statistics from content and style images. 
AdaIN~\cite{AdaIN} proposes to match the mean and variance of intermediate features between content and reference images. 
Other works use higher-order statistic analysis~\cite{style_5,Deepphoto,LT,style_4}. 
For instance, \cite{Deepphoto} explores the non-local feature correlations, whereas WCT~\cite{WCT} refines style transfer by directly embedding the Whitening and Coloring Transforms into a network.  
Recent works tackle it with Generative Adversarial Networks (GAN)~\cite{GAN_3,GAN_7,GAN_8,GAN_9}. A representative work is CycleGAN~\cite{cycleGAN} that uses two sets of GANs to form a mapping loop between domain A and B. 
\cite{GAN_3} proposes to disentangle contents and styles from images so the stylization can be resolved in the style space, while \cite{GAN_14} includes segmentation information for conditional editing.

\paragraph{Multiple style transfer } is a less explored topic. 
The goal is to mix multiple styles and add them to the content image. Most  approaches~\cite{style_1,AdaIN,multi_1,multi_2,Deepphoto,Ulyanov} cast it as feature interpolation, \ie interpolate between different styles and transfer the new style to the content images. \cite{AdaIN} interpolates the means and variances of the style feature maps, whereas \cite{multi_3} adds the style interpolation into the loss and trains a model for different style mixture. 
However, these solutions are time-consuming and the range of mean and variance for different style images varies a lot, and hence simple interpolation does not guarantee the desired style mixture. 
Other works use GANs, \eg \cite{multi_2} uses 1D codes as a condition for style transfer; however, the network is trained on a fixed number of styles, and hence it cannot transfer styles outside the training data. 
\cite{multi_1} proposes a GAN to train a spatial transformation matrix; however, the network performs poorly on styles outside the training data, and there are no multiple or convincing style transfer studies.

\section{Method}
Here, we describe the Variational AutoEncoder for latent space-based Style Transfer (\net) that performs multiple style transfer by projecting nonlinear styles to a linear latent space, and it fuses them by linear interpolation (Figure~\ref{fig:Figure1}). 
It consists of an Image AutoEncoder (\textcolor{vblue}{\textbf{IAE}}), \ie an encoder-decoder performing image reconstruction (Section~\ref{sub:iae}); and a Variational autoencoder-based Linear Transformation (\textcolor{vred}{\textbf{\subnet}}), 
responsible for latent space-based style manipulation (Section~\ref{sub:vaelt}). 
The training comprises 
two phases (Figure~\ref{fig:Figure1}(a)).  
First, IAE is trained for image reconstruction. 
Second, IAE is frozen and 
\subnet learns the covariance of different styles via the \textcolor{vyellow}{\textbf{linear-transformation module}} and then the \textcolor{vmagenta}{\textbf{variation module}} projects them to different components of a multi-dimensional Gaussian distribution via KL minimization. 
At test time, the style transfer is processed in the latent space via a mixture of Gaussian distributions (\textcolor{vmagenta}{\textbf{magenta}} in Figure~\ref{fig:Figure1}(b)).

\subsection{Image AutoEncoder (IAE)}
\label{sub:iae}
\textcolor{vblue}{\textbf{IAE}} is a symmetric encoder-decoder that extracts features for image reconstruction. 
The encoder's structure follows VGG-19~\cite{VGG} by keeping all conv layers and discarding the fc ones. 
The decoder is symmetric to the encoder and up-samples the feature abstraction to reconstruct the input image. 
Unlike~\cite{cycleGAN,u-net,RefVAE}, IAE has no short connections. 
IAE performs style transfer in the feature domain. 
The encoder learns a one-to-one mapping \textbf{E} from images \textbf{X} to features \textbf{M}, while the decoder learns a mapping \textbf{D} to reconstruct the images. 
These mappings ensure that each image corresponds to a unique and compact feature.

\begin{figure*}[t]
	\begin{center}
		\centerline{\includegraphics[width=\textwidth]{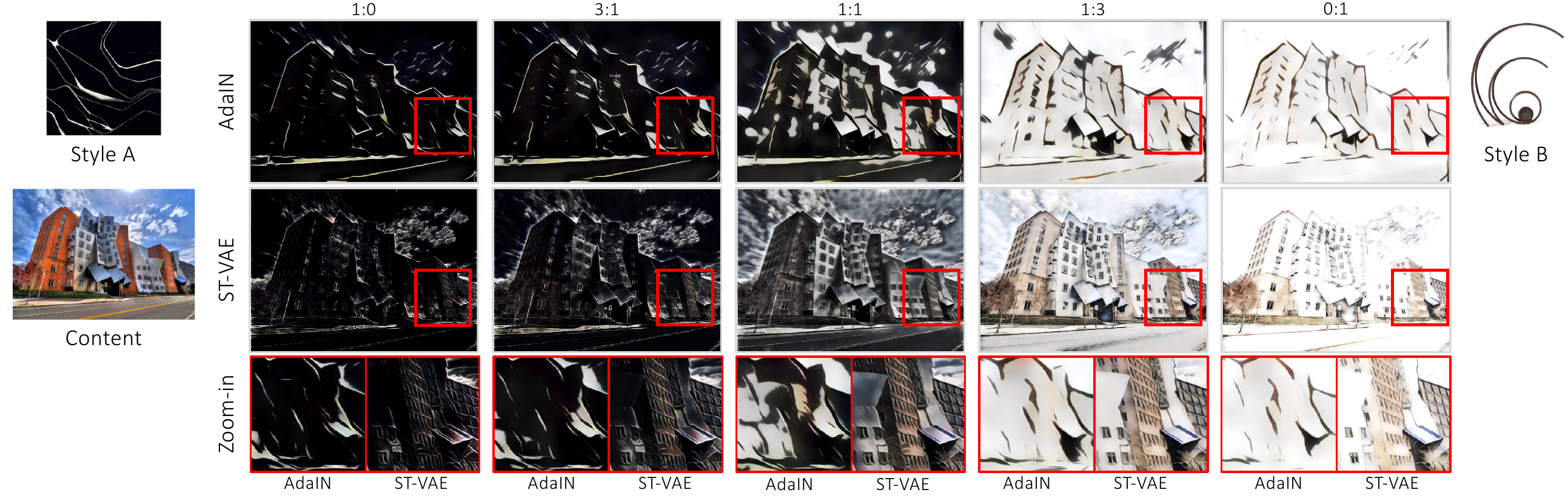}}
		\caption{\small{Image synthesis using two style images for the same content image for AdaIN~\cite{AdaIN} and \net (ours). The weights of the styles  are shown above each stylized image. \net transitions between styles smoother than AdaIN, especially on the building edges and windows.}}
		\label{fig:Figure3}
	\end{center}
\end{figure*}

\subsection{VAE-based Linear Transformation (\subnet)}
\label{sub:vaelt}
\textcolor{vred}{\textbf{\subnet}} is built upon a Variational AutoEncoder and is responsible for latent space based style manipulation. 
It consists of a  \textcolor{vyellow}{\textbf{linear transformation}} (Section~\ref{subsub:lt}), which learns the covariance matrices of the content and style features; 
and a \textcolor{vmagenta}{\textbf{variation module}} (Section~\ref{subsub:vae}), which projects the styles to a normal Gaussian space. 
After, the content and style covariance matrices are multiplied for stylization.

\paragraph{Notation.}
Let ${F_c,F_s}\in\mathbb{R}^{C \times N}$ be the vectorized feature maps of content and style image, respectively that are obtained at the top-most encoder layer. $\mathit{N}$ is the feature length for content and style image and \textit{C} is the number of channels. 

\subsubsection{Linear Transformation module}
\label{subsub:lt}
To perform content and style transfer, we deploy a two-fold objective. 
First, we find a linear transformation $\mathit{T}\in\mathbb{R}^{C \times C}$ to transfer content ${F_c}$ to the desired feature maps  $F_d$, 
such that $F_d=TF_c$. 
Second, we find a nonlinear mapping model $\phi$, such that $\phi_s=\phi(F_s)$, where $\phi_s$ is the transformed style feature maps. 
This is in line with~\cite{WCT,LT} that cast the stylization as a covariance matching process. 
Our objective is:
\begin{small}
	\begin{equation} \tag*{(1)}
	\begin{matrix}
	\begin{split}
	& \mathit{L_{style}}=\frac{1}{NC}||\bar{F}_d\bar{F}_d^T-\bar{\phi}_s \bar{\phi}_s^T||^\mathit{l} \\
	& s.t. \bar{F}_c=F_c-\mathit{mean}(F_c), \bar{F}_d=T\bar{F}_c. 
	\label{eq:Equation1}
	\end{split}
	\end{matrix} 
	\end{equation}
\end{small}
Equation~\ref{eq:Equation1} describes the \textit{l}-th order minimization, where $\bar{F}_d$ is the desired feature vector. 
$\bar{F}_d\bar{F}_d^T$ is the covariance of $\bar{F}_d$. Following~\cite{WCT}, we find the covariance of $\bar{F}_d$, with the whitening process, \ie singular value decomposition. We find the eigenvector matrix E and the diagonal eigenvalue matrix D as: $cov(\phi_s)=\bar{\phi}_s \bar{\phi}_s^T=E_sD_sE_s^T$, 
and hence we estimate the covariance as: $cov(\bar{F}_d)=T\bar{F}_c \bar{F}_c^TT^T=TE_cD_cE_c^TT^T$. 
Therefore, the generalized transformation is: 

\begin{table*}[ht!]
\begin{minipage}[b]{0.71\textwidth}
\centering
\includegraphics[width=1\textwidth]{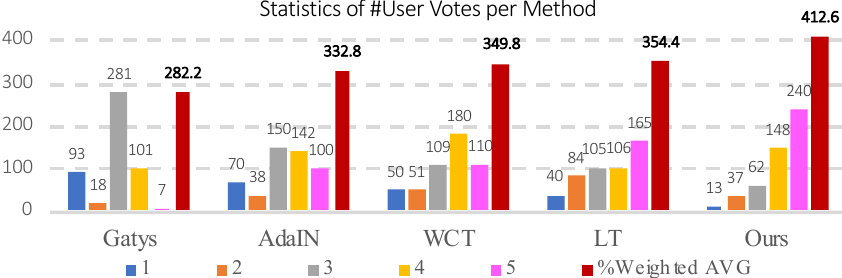}
\captionsetup{skip=0cm, width=0.99\textwidth}
    \captionof{figure}{\small{Statistics of \% user scores on style transfer. Different colors represent different scores from 1 to 5 (5 being the highest). 20 users were asked to rate synthesized images.  Overall, we collect 500 votes for each method and observe that images from \net are unanimously rated the best.}}
    \label{fig:Figure4}
\end{minipage}
\hfill
\begin{minipage}[b]{0.27\textwidth}
\footnotesize{
\begin{tabular}{lrrr}
\hline
					Image & 256$\times$   & 512$\times$   & 1024$\times$   \\ 
					size & 256~~ & 512~~ & 1024~~ \\
					\hline
					Gatys~\cite{Gatys}      & 16.51 & 59.45 & 251.44 \\ 
					AdaIN~\cite{AdaIN}      & 0.019 & 0.071 & 0.288  \\
					WCT~\cite{WCT}        & 0.922 & 1.080 & 4.001  \\
					LT~\cite{LT}         & {\color{red}0.01}  & {\color{red}0.036} & {\color{blue}0.146}  \\
					Ours       & {\color{red}0.01}  & {\color{blue}0.041} & {\color{red}0.145} \\ \hline
\end{tabular}}
\captionof{table}{\small{Running time comparison on different image resolutions, measured in seconds using the original source code on a GTX 1080Ti GPU. {\color{red}Red} indicates the best results, {\color{blue}blue} indicates the second best results.}}
\label{tab:Table1}
\end{minipage}
\vspace{2mm}
\end{table*}

\begin{footnotesize}
	\begin{equation} \tag*{(2)}
	\begin{matrix}
	\begin{split}
	& TE_cD_cE_c^TT^T=E_sD_sE_s^T \\
	& T=\big(E_sD_sE_s^T\big) \Sigma \big(E_cD_cE_c^T\big)\\
	& T=T_1\big(E_sD_sE_s^T\big) \times T_2\big(E_cD_cE_c^T\big),  
	\label{eq:Equation2} 
	\end{split}
	\end{matrix} 
	\end{equation}
\end{footnotesize}

\noindent  where $\mathit{\Sigma}\in\mathbb{R}^{C \times C}$ is \textit{C}-dimensional orthogonal matrix. 
Equation~\ref{eq:Equation2} shows that the transformation \textit{T} is determined by the covariance of the content and style image features. 
Once \textit{T} is calculated, $F_d$ is obtained by $T\bar{F}_c+\mathit{mean}(F_s)$, which aligns it to the mean and covariance of the style image.

\subsubsection{Variation module}
\label{subsub:vae}

The \textcolor{vmagenta}{\textbf{variation module}} is a projection model responsible for multiple style transfer. Only a few works explore controllable or weighted style transfer~\cite{style_1,AdaIN,style_9} by manually adjusting the weights for linear interpolation, but they cause inconsistent style transition as the features are nonlinear. Instead, we embed the variation module into the transformation model to map features into a linear space spanned by mixture models.

VAE is defined as: P($\mathbf{X}$)=$\int P(\mathbf{X},z)P(z|\mathbf{X})\, dz$, where \textbf{X} is the input and \textit{z} is sampled from the latent space \textbf{Z}~\cite{RefVAE,VAE_1,SRVAE}. To regularize the latent space, we use the Kullback–Leibler (KL) divergence that measures the probability close to a normal distribution. The variation module learns parameters $\theta$ for maximizing the data log likelihood $P_{\theta}(\mathbf{X})$:

{
\begin{footnotesize}
	\begin{equation} \tag*{(3)}
	logP_{\theta}(\mathbf{X})=E_{Q_{\omega}(z|X)}[logP_{\theta}(\mathbf{X},z)] 
	-KL[Q_{\omega}(z|\mathbf{X})||P_{\theta}(z|\mathbf{X})].
	\label{eq:Equation3}
	\end{equation}
\end{footnotesize}
}
Equation~\ref{eq:Equation3} shows that the encoder learns parameters $\omega$ to approximate posterior $Q_{\omega}(z|\mathbf{X})$, while the decoder learns $\theta$ to represent the likelihood $P_\theta(\mathbf{X},z)$. 
The real prior distribution $P_{\theta}(z|\mathbf{X})$ is a Gaussian distribution and the approximated posterior follows $z\sim Q_{\omega}(z|\mathit{x}_i)=\mathit{N}(z;\mu_i,\sigma^2_i\mathit{I})$. 
We use the variation module to cast the multiple style transfer problem as a generative sampling process: by projecting arbitrary style images to a hidden distribution, each style image corresponds to one sample on the latent space. Thus, the multiple style transfer becomes data interpolation in the latent space, performed by a multivariate Gaussian mixture model.

\paragraph{Training Loss. }
We train our model using the style $\mathit{L}_{\text{style}}$ and content losses $\mathit{L}_{\text{content}}$ and KL divergence as follows: 

\begin{footnotesize}
	\begin{equation} \tag*{(4)}
	\begin{matrix}
	\begin{split}
	& \mathit{L}_{\text{content}}=\mathbb{E}||\psi(T(\mathbf{X}))-\psi(\mathbf{X})||_1, \text{and} \\
	&\mathit{L}_{\text{\subnet}}=\mathit{L}_{\text{content}}+\lambda\mathit{L}_{\text{style}}+\beta KL[Q_{\phi}(z|\mathbf{X})||\mathit{N}(0,1)],\\
	\end{split}
	\end{matrix} 
	\label{eq:Equation4}
	\end{equation}
\end{footnotesize}

\noindent where $\lambda$ and $\beta$ are the weighting parameters to balance style and KL losses, $\psi_i$ is the \textit{i}-th feature maps extracted from the pre-trained VGG-16 model~\cite{VGG}. Recall that we use VGG-16 to compute losses, while we modify VGG-19 for IAE.

\begin{figure}[ht!]
	\begin{center}
		\centerline{\includegraphics[width=0.96\linewidth]{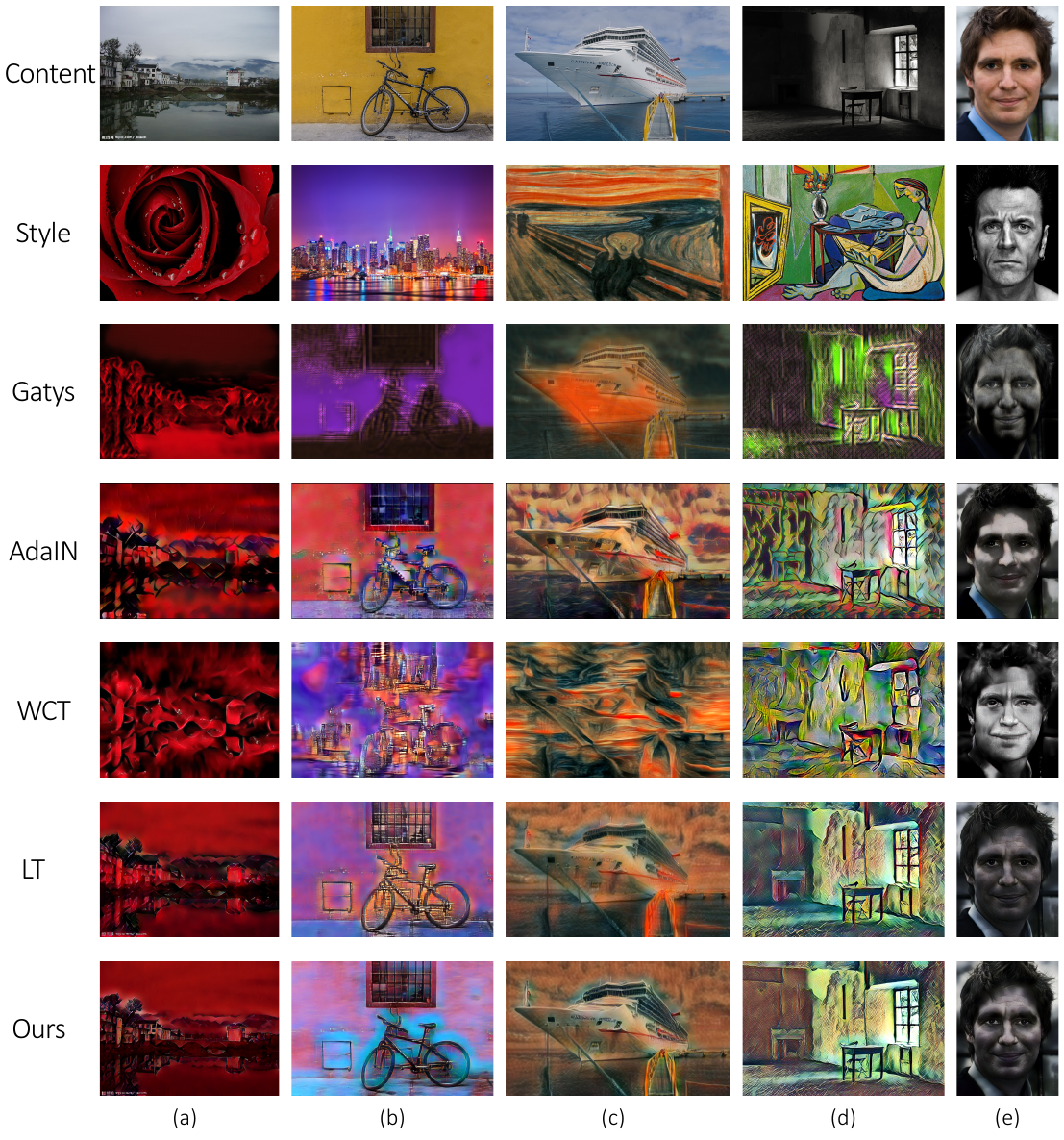}}
		\caption{\small{Results of seven style transfer methods. \net has more similar affinity with the content than other methods, \eg it preserves the  objects in (b,d), while others fail to produce detailed structures.}}
		\label{fig:Figure2}
	\end{center}
\end{figure}

\section{Experiments}
\label{sec:experiments}
We report results on single and multiple style transfer, runtime and quantitative evaluations (more examples in sup.\ material).

\paragraph{Implementation details.} 
For content loss, we use \textit{relu4\_1} to compute differences between content and stylized images. 
For perceptual loss, we combine first- and second-order statistics to measure the similarity between reference and stylized images. 
We train IAE on COCO~\cite{COCO}.  
For \subnet, we use COCO as content and WikiArt~\cite{WikiArt} as style. 
At training, we keep the ratio and crop a region of 256$\times$256 as patches. For augmentation, we randomly flip and rotate the contents. 
We train with Adam optimizer, a learning rate of $10^{-4}$ and batch size of 8 for 100k iterations (8h, NVIDIA GTX1080Ti GPU).

\paragraph{Discussion.}
\net transfers textures and styles efficiently (appr.\ 100 fps). 
For efficiency, the input features are first compressed for transformation learning and then uncompressed to match the original dimension (`conv' in Figure~\ref{fig:Figure1}). 
Instead of fixed-size content and style~\cite{Gatys,Deepphoto}, \net does not depend on the image resolution, thus handling arbitrary style transfer.

\paragraph{General style transfer.} 
\net performs robust style transfer without affecting the structure of the content images. To show its effectiveness, we compare it to the state of the art: Gatys~\cite{Gatys}, AdaIN~\cite{AdaIN}, WCT~\cite{WCT}, and LT~\cite{LT}. 
Figure~\ref{fig:Figure2} shows five content and style images and the results with all methods. 
For a fair evaluation, we choose content images from~\cite{Deepphoto} and~\cite{ABPN} that are not part of our training set. 
\net successfully transfers the desired styles and textures and preserves the details of the content better. For instance, in Figure~\ref{fig:Figure2}(a) \net clearly reconstructs the windows and doors.

\paragraph{Multiple style transfer.} 
Figure~\ref{fig:Figure3} shows the style mixture results when using two style images for \net and AdaIN~\cite{AdaIN}. Note, the style interpolation is done by assigning different weights to the style images. We focus on the texture transition. 
We observe that \net successfully preserves the content information better than AdaIN while transitioning between styles in a smoother way. 
For instance, it successfully preserves the clouds and windows while transferring styles, while AdaIN losses these details. Furthermore, \net results in smooth changes between foreground and background without any boundary effects, while AdaIN fails.

\paragraph{Quantitative evaluation.}
To evaluate \net, we conduct a user study, where users are presented with 5 synthetic images in random order and are asked to rate their quality from 1 to 5 (5 the highest). 
We use five methods: Gatys, AdaIN, WCT, LT and ours. 
For each method, we synthesize 400 images (10 content, 40 styles) and randomly select 20. 
We collect 500 votes from 20 users and report the results in Figure~\ref{fig:Figure4}, where we observe that \net is favoured amongst all.

\paragraph{Computational cost.} 
Table~\ref{tab:Table1} reports the run-times on style transfer with different resolutions. 
In most cases, \net leads to the lowest run-time; for instance, for low-resolution images it generates images within 0.01s.

%-------------------------------------------------------------------------
\section{Conclusion}
We introduced \net, a Variational AutoEncoder based style transfer method that maps features into a multivariate Gaussian distribution for both single and multiple style transfer with more consistent style transitions. The linear transformation enables feed-forward training and testing, thus making \net very efficient. 
Our experiments show that \net performs favourably against the state of the art, both quantitatively and qualitatively. 
Future work involves extending it to videos by exploiting the temporal continuity of frames~\cite{kalogeiton2017action}.

\noindent \textbf{Acknowledgements.} This work was partly funded by the Google chair at École Polytechnique.

{\small
\bibliographystyle{IEEEbib}
\bibliography{shortstrings,egbib}
}

\end{document}